\documentclass[sigconf]{acmart}

\usepackage{balance}

\def\BibTeX{{\rm B\kern-.05em{\sc i\kern-.025em b}\kern-.08emT\kern-.1667em\lower.7ex\hbox{E}\kern-.125emX}}
    
\newcommand{\figref}[1]{Figure~\ref{#1}}

\newcommand{\tabref}[1]{Table~\ref{#1}}
\newcommand{\eqnref}[1]{Equation~(\ref{#1})}

\newcommand{\etal}{{\it et al.}}
\newcommand{\eg}{{\it e.g.}}
\newcommand{\ie}{{\it i.e.}}

\copyrightyear{2019} 
\acmYear{2019} 
\acmConference[MM '19]{Proceedings of the 27th ACM International Conference on Multimedia}{October 21--25, 2019}{Nice, France}
\acmBooktitle{Proceedings of the 27th ACM International Conference on Multimedia (MM '19), October 21--25, 2019, Nice, France}
\acmPrice{15.00}
\acmDOI{10.1145/3343031.3350864}
\acmISBN{978-1-4503-6889-6/19/10}

\begin{document}

\fancyhead{}

\title{Preserving Semantic and Temporal Consistency for Unpaired Video-to-Video Translation}


\author{Kwanyong Park$^1$, Sanghyun Woo$^1$, Dahun Kim$^1$, Donghyeon Cho$^2$, In So Kweon$^1$}

\affiliation{%
  \institution{\textsuperscript{1}Korea Advanced Institute of Science and Technology(KAIST), \textsuperscript{2}Chungnam National University}
}
\email{{pkyong7,shwoo93}@kaist.ac.kr,{mcahny01,cdh12242}@gmail.com,{iskweon77}@kaist.ac.kr}

\renewcommand{\shortauthors}{Park, et al.}

%
\begin{abstract}
In this paper, we investigate the problem of unpaired video-to-video translation. Given a video in the source domain, we aim to learn the conditional distribution of the corresponding video in the target domain, without seeing any pairs of corresponding videos. While significant progress has been made in the unpaired translation of images, directly applying these methods to an input video leads to low visual quality due to the additional time dimension. In particular, previous methods suffer from semantic inconsistency (i.e., semantic label flipping) and temporal flickering artifacts. To alleviate these issues, we propose a new framework that is composed of carefully-designed generators and discriminators, coupled with two core objective functions: 1) content preserving loss and 2) temporal consistency loss. Extensive qualitative and quantitative evaluations demonstrate the superior performance of the proposed method against previous approaches. We further apply our framework to a domain adaptation task and achieve favorable results.
\end{abstract}


%

%
\begin{teaserfigure}
\begin{tabular}{@{}c@{}}
\includegraphics[width=0.99\linewidth]{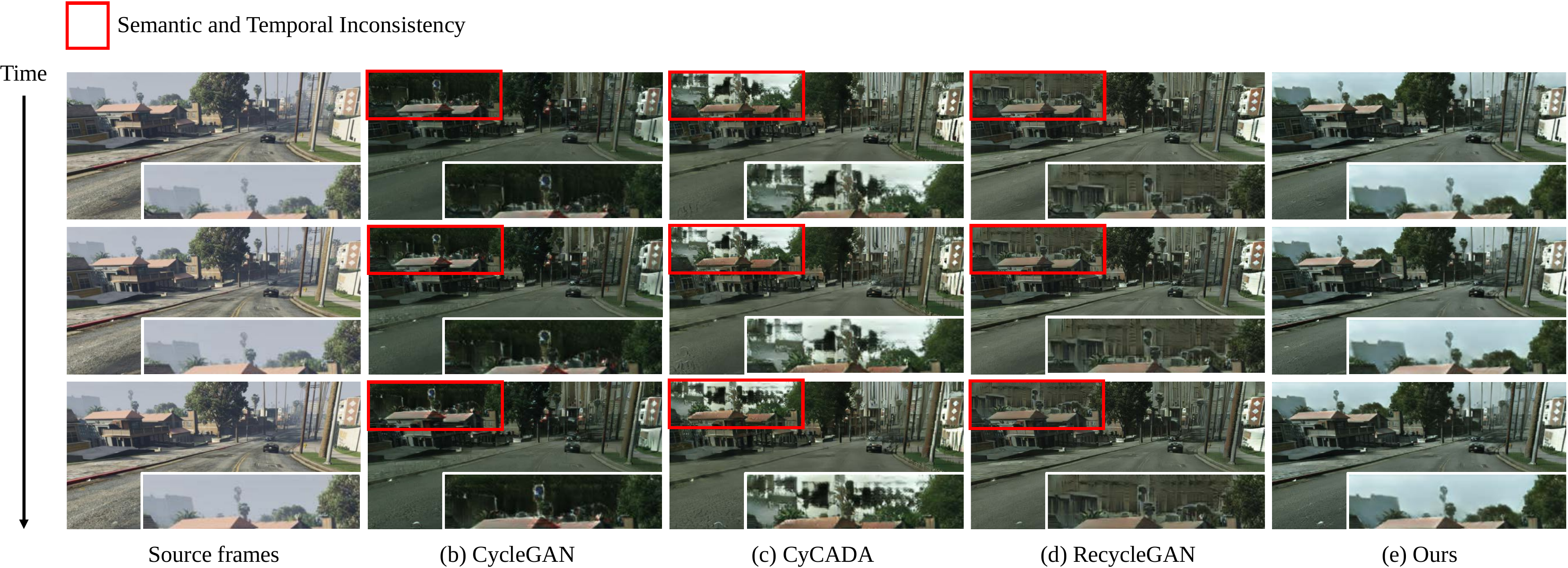} \\
\end{tabular}
\vspace{-3mm}
\caption{ \textbf{Video translation results (VIPER $\rightarrow$ Cityscapes).} We empirically observe that previous state-of-the-art methods suffer from two main issues: 1) semantic label inconsistency and 2) temporal inconsistency. The proposed framework is robust to semantic label flipping and temporal flickering artifacts. \textit{Best viewed in color}. }
\label{fig:teaser}
\end{teaserfigure}

%
\maketitle

\section{Introduction}
Cross-domain video-to-video translation has a wide range of applications in computer vision, robotics, and graphics tasks. This technology is particularly effective for modeling visual domains where capturing or labeling real-world data is expensive, e.g., when creating the visual dynamics for human demonstrations that teach robots, or for labeling self-driving scenes with lower amounts of real data. When direct mapping between data points in the source and target domains is available, cross-domain translation can be solved by a simple regression~\cite{isola2017image} or a conditional generative formulation~\cite{isola2017image}. However, creating such paired (\ie~aligned) datasets is neither practical nor scalable in real-world scenarios. In this paper, we deal with a more practical and challenging setting where such direct supervision is not available. We aim to translate a video from a synthetic domain to a real-world domain in the absence of a paired training dataset: \textit{unpaired video-to-video translation}.

Despite the great amount of progress that has been made on the unpaired image translation problem~\cite{liu2017unsupervised,zhu2017unpaired,hoffman2017cycada,huang2018multimodal}, it is still challenging to extend these methods to its video counterpart. A straightforward way to perform unpaired video-to-video translation is to apply a per-image translation to each frame. However, this naive approach has several limitations. First, the 2D based constraints prevent the use of video dynamics, which is a valuable training signal for video translation models~\cite{bansal2018recycle}.
Second, it inevitably generates temporal inconsistencies and causes severe flickering artifacts. 
In general, original/new content should not abruptly disappear/appear in the video (\ie, flickering artifacts). However, previous approaches clearly suffer from this problem, as shown in \figref{fig:teaser} (b) and (c); the contents significantly varies over time.

Recently, there have been some attempts to utilize temporal constraints in generating videos. Lai~\etal~\cite{lai2018learning} designed a deep network to post-process per-frame generated videos to temporally smooth videos. We build our baseline method by combining this post-processing method with a state-of-the-art per-frame translation method. Experiments show that our proposed method outperforms this strong baseline. Bansal~\etal~\cite{bansal2018recycle} proposed the RecycleGAN framework for unpaired video translation. While sharing a similar goal, we empirically observed their limitations. First, their method frequently fails to maintain the semantic of the scene parts before and after translation. For example, it sometimes translates the pixels in the \textit{sky} region into the \textit{building} structure when translating between VIPER and Cityscapes videos (see \figref{fig:teaser} (d)). Another limitation is the temporally unstable and blurry video results.

In this work, we introduce a new framework for unpaired video-to-video translation. Our goal is to learn video translation between two unpaired domains, such that the video results are semantically consistent with the original content, as well as temporally smooth. Our generator is trained with two complementary objective functions: 1) content preserving loss and 2) temporal consistency loss. Following the generator is a fusion block where the pixels that are newly generated and the pixels that are warped from the previous output frame are adaptively combined to produce the final output frame. Intuitively, re-using the warped pixels leads to more stable video results than generating all pixels from scratch for every frame (e.g., as in RecycleGAN~\cite{bansal2018recycle}). The fusion algorithm is trained using both content preserving loss and temporal consistency loss. Jointly using these losses not only enforces semantic and temporal consistency, but it also compensates the intrinsic error in predicted flow (e.g., flow from the FlowNet~\cite{ilg2017flownet}). 
By using translated video results that are semantically and temporally consistent, our framework greatly boosts the semantic segmentation performances in domain adaptation setting as well.

Our main contributions can be summarized as follows:
\begin{itemize}
\item  We propose a unified framework for unpaired video-to-video translation. The proposed model is composed of carefully-designed generators and discriminators that are coupled with two core objective functions: 1) content preserving and 2) temporal consistent generation.

\item To the best of our knowledge, this is the first work that introduces optical flow into unsupervised video translation.

\item  We conduct extensive qualitative and quantitative evaluations and show the efficacy of our framework. Moreover, we show our framework produces favorable results on a domain adaptation task.

\end{itemize}

\begin{figure*}[t]
\begin{tabular}{@{}c@{}}
\includegraphics[width=0.96\linewidth]{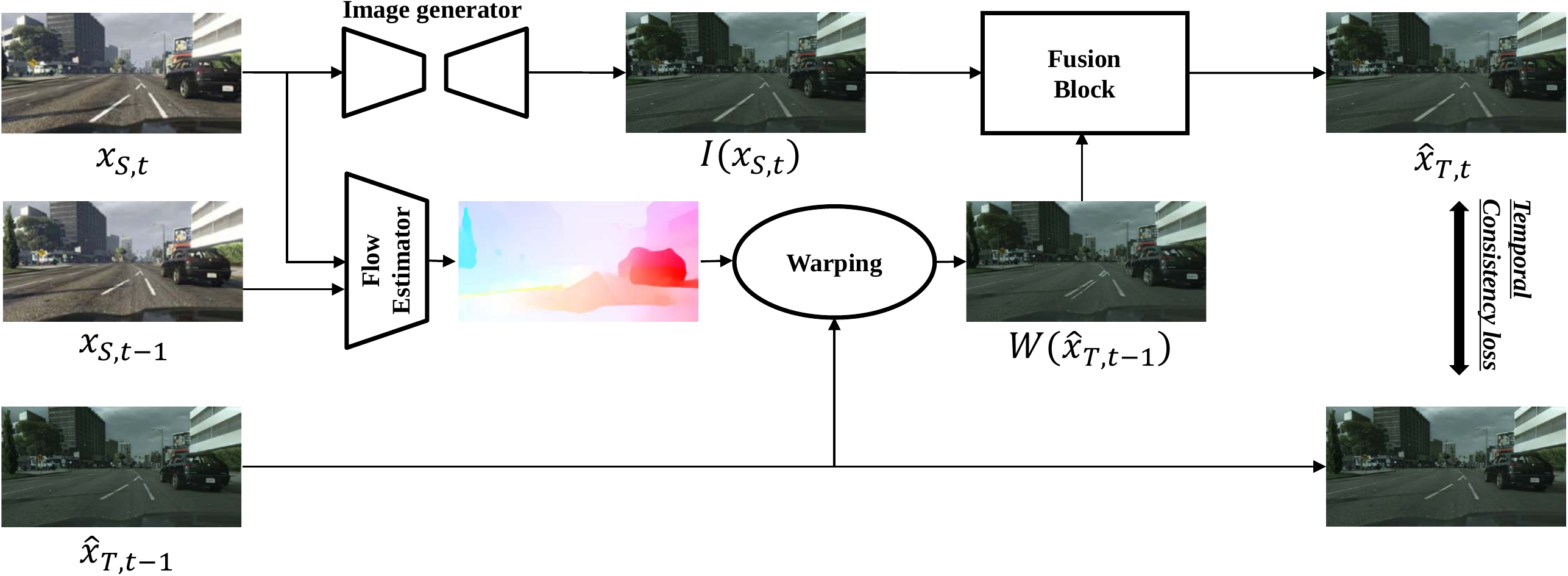} \\
\end{tabular}
\caption{ \textbf{Overview of recurrent generator.} Our network takes in current, previous source frames (${x}_{S,t}, {x}_{S,t-1}$) and the previously generated frame ($\hat{x}_{T,t-1}$) as an input. The generator then translates input frames to target-like source frame ($\hat{x}_{T,t}$). We employ recurrent feedback (\ie~flow estimator and fusion block) to blend the warped pixels and the newly synthesized pixels smoothly. }
\label{fig:generator}
\end{figure*}

\section{Related Works}

\subsection{Generative Adversarial Networks (GANs)}
In GAN~\cite{goodfellow2014generative} training, a generator and a discriminator play a zero-sum game. The generator is trained to fool the discriminator and produces realistic synthetic data so that the discriminator cannot distinguish between the synthesized and real data. Previous works have successfully applied GAN framework to various applications such as image generation~\cite{denton2015deep,radford2015unsupervised,zhao2016energy}, image editing~\cite{zhu2016generative,shetty2018eiting} and feature learning~\cite{salimans2016improved,donahue2016adversarial}. A variety of works, in fact, have adopted the conditional GAN framework~\cite{isola2017image} for these image-to-image translation problems. Recent approaches have been proposed to deal with the video translation problem~\cite{wang2018video}, but they require the training data of input-output video pairs, which is extremely expensive or unavailable in most real-world scenarios. Lines of work also exist with unpaired settings, where the training pairs are not given~\cite{yoo2016pixel,taigman2016unsupervised,shrivastava2017learning,bousmalis2017unsupervised,donahue2016adversarial,dumoulin2016adversarially,liu2017unsupervised,zhu2017unpaired}. Liu~\etal~\cite{liu2017unsupervised} proposed the UNIT framework, which assumes a shared latent space such that image pairs in two domains are mapped to the same latent code. Zhu~\etal~\cite{zhu2017unpaired} suggested the use of the cycle-consistency constraint in an adversarial learning framework. This enforces the inverse mapping from the target domain to the source domain to produce an image identical to the original source image. The authors show that the cycle-consistency constraint can achieve compelling translation results without expensive manual labeling (~\ie, unpaired data). Variants of this CycleGAN have been proposed in the spatial domain~\cite{hoffman2017cycada}, but they consider only the spatial information while ignoring the temporal dynamics. Bansal~\etal~\cite{bansal2018recycle} proposed the ReCycleGAN framework that utilizes both spatial and temporal constraints. However, we observe semantic inconsistency (i.e., semantic label flipping) in their translation results. Based on the cycle consistency method~\cite{zhu2017unpaired}, we take an important next step in video translation by overcoming the two main limitations of the previous methods: (1) semantically incorrect translation, and (2) temporal inconsistency.

\subsection{Video Synthesis}

The use of GAN also provides a way of synthesizing videos and temporal information. TGAN~\cite{saito2017temporal} used a temporal generator and an image generator that generates a set of latent variables and image sequences, respectively. Similarly, MoCoGAN~\cite{tulyakov2018mocogan} decomposed the latent space to motion and content and used a recurrent neural network to generate a sequence of motion codes. These methods often produce short-length and low-resolution videos.

Given a reference painting, video style transfer models~\cite{chen2017coherent} are trained to transfer the reference style into an input video. 
The video prediction task aims to predict future frames conditioning on the given frames, which is also related to our task. Many of these models~\cite{Lotter2017prediction,Finn2016prediction} are trained with simple L1/L2 reconstruction losses. Hence, they fail to produce long-duration videos and often generate blurry videos due to the regress-to-the-mean problem. Our approach is different in that we do not attempt to predict future camera or object motions.

There are some special cases which can also be considered as the video synthesis problem such as video super-resolution~\cite{Sajjadi2018videoSR,huang2018videoSR}, video manipulating~\cite{lee2019instert}, video decaptioning~\cite{kim2019deep1}, and video inpainting~\cite{wang2018videoinp,kim2019deep2}. However, these methods depend on problem-specific designs and constraints, and thus do not likely generalize to other cross-domain translation problems.

\section{Method}

\sloppy Given a source set $X_{S}$ consisting of a temporally ordered frame sequence $x_{S} := \{x_{S,1},x_{S,2},...,x_{S,t},...\}$ $(x_{S} \in X_{S})$ and a target set $X_{T}$ consisting of  $x_{T} := \{x_{T,1},x_{T,2},...,x_{T,t},...\}$ $(x_{T} \in X_{T})$, we consider an unpaired setting where no information is provided as to which $x_{S}$ matches which $x_{T}$. The goal is to learn a mapping between videos of the two domains (~\ie, the source and target). While existing per-frame approaches do not use temporal information, we take advantage of the temporal flow in a video sequence. The proposed framework and learning objectives are detailed below.

\begin{figure*}[t]
\begin{tabular}{@{}c@{}}
\includegraphics[width=0.96\linewidth]{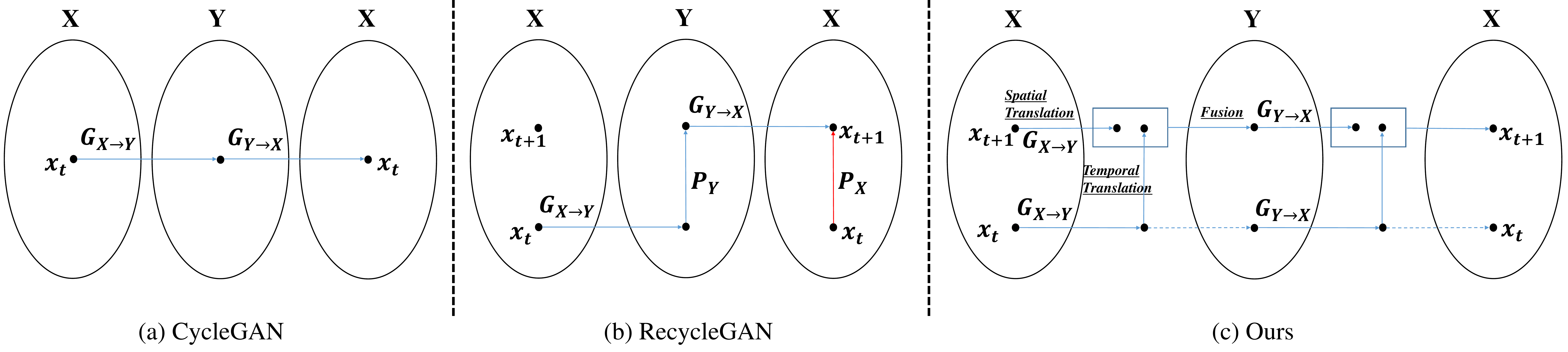} \\
\end{tabular}
\caption{ \textbf{Different cycle-consistency constraints.} (a) \textbf{CycleGAN}:~\cite{zhu2017unpaired} proposed the use of cycle-consistency loss to deal with the problem of unpaired data setting. The approach however only considers independent 2D images. (b) \textbf{RecycleGAN}:~\cite{bansal2018recycle} firstly proposed the use of temporal cues in videos. For video generation, however, the approach only considers one dimension, \eg ~$G_{S\rightarrow T}$ or $P_Y$, which is ineffective. Moreover, because the approach requires a learnable temporal predictor, $P$, this imposes an optimization difficulty. (c) Instead of using a future frame predictor, we use optical flow to warp the previously generated frame and combine it with the currently translated frame softly. Such an approach not only produces videos of better visual quality but also effectively imposes a spatio-temporal constraint.}
\label{fig:mapping}
\end{figure*}

\subsection{Recurrent Generator}

We design a recurrent generator $G_{S\rightarrow T}$ that sequentially translates input video frames from a source domain,~\ie $x_{S}$, to a target domain,~\ie $\hat{x_{T}}$. The overview of the proposed generator is shown in \figref{fig:generator}. This model $G$ consists of three sub-modules: an image generator $I_{S \rightarrow T}$, a flow estimator, and a fusion block. We employ the markovian assumption so that the generation of the $t$-th frame $\hat{x}_{T,t}$ depends only on 1) the current input frame ${x}_{S,t}$, 2) the previous input frame ${x}_{S,t-1}$, and 3) the previous output frame $\hat{x}_{T,t-1}$.

\begin{equation}
\begin{split}
\hat{x}_{T,t} = G_{S\rightarrow T}({x}_{S,t}, {x}_{S,t-1}, \hat{x}_{T,t-1}).
\end{split}
\label{eqn:generator}
\end{equation}

The whole video output $\hat{x_{T}}$ is obtained by applying $G_{S \rightarrow T}(\cdot)$ sequentially in an auto-regressive manner. For each time step, the image generator $I_{S \rightarrow T}$ takes a current source frame $x_{S,t}$ as input and outputs a translated frame $I({x}_{S,t})$ which we consider as an intermediate result.

\sloppy In order to capitalize on temporal flow in the videos, we design our framework to connect between different time steps. For each time step, a flow estimator computes the optical flow between two consecutive frames ${x}_{S,t-1}$, and ${x}_{S,t}$. The computed flow field is used to warp the previous output frame $\hat{x}_{T,t-1}$ onto the current time step. As a flow estimator, we adopt FlowNet2~\cite{ilg2017flownet} which is pre-trained on the synthetic FlyingChairs and FlyingThings 3D datasets~\cite{DFIB15,mayer2016large}.

Although this stage of temporal warping borrows traceable pixels from the previous output frames, there are still pixel regions that have to remain untouched. In particular, the flow estimation is not reliable for pixels at occlusions or newly appeared scene parts, so these pixels have to be newly synthesized by the image generator. To combine these two cases in a learnable manner,  we introduce a fusion block. 
Our fusion block learns to regress this soft mask which adaptively blends the warped pixels and generated pixels into \textit{one frame}.
In other words, the fusion block($F$) computes a one-channel soft fusion mask($m$) to combine the previous warped frame $W(\hat{x}_{T,t-1})$ and the current synthesized frame $I(\hat{x}_{S,t})$. The block takes the absolute difference between the two frames and outputs the fusion mask. This can be formulated as:

\begin{equation}
\begin{split}
m &= F(I(\hat{x}_{S,t})-W(\hat{x}_{T,t-1})), \\
\hat{x}_{T,t} &= m.*W(\hat{x}_{T,t-1})+(1-m).*I(\hat{x}_{S,t}), \\
\end{split}
\end{equation}
where .* is an element-wise multiplication operator. The value of the mask varies between 0 and 1. For the pixels that are warped from the previous frame are encouraged to be reused if they are traceable and reliable (the mask value tends to be 1). This temporal warping supports temporal coherence by connecting pixels over time steps. On the other hand, the remaining pixels are encouraged to take their value from the newly synthesized frame $I(x_{S,t})$ (the mask value tends to be 0).

\subsection{GAN Loss}
We apply an adversarial learning strategy~\cite{goodfellow2014generative} to train our model  $G_{S\rightarrow T}$ (or $I_{S\rightarrow T}$). We note that the GAN loss for $I$ is omitted for clarity.
We train the generator $G_{S\rightarrow T}$ with a discriminator $D_{T}$ that is adversarially trained to distinguish between real samples of ${x}_T$ and generated samples $\hat{x}_T$. The corresponding loss function is 
\begin{equation}
\begin{split}
\min_{G} \max_{D} L_{gan,forward}(G,D) &= _{x_T \sim X_T} [log D(x_{T,t})] \\
                                       + _{x_S \sim X_S} [log(1-D(&G({x}_{S,t}, {x}_{S,t-1}, \hat{x}_{T,t-1})))]. 
\end{split}
\label{eqn:gan_loss}
\end{equation}

Note that the objective function does not require paired data and only requires access to samples from source domain \{$X_S$\} and target domain \{$X_T$\}. Different subscripts $\{S\}$ and $\{T\}$ are used to emphasize the unpaired setting. This objective function encourages the generated samples $G_{S \rightarrow T}(\cdot)$ to be indistinguishable to data distribution drawn from $X_T$.

\subsection{Spatio-temporal Cycle-consistency Loss}

We employ a cycle-consistency constraint in our method ~\cite{zhu2017unpaired}. First, we introduce another generator that maps from target to source, $G_{T \rightarrow S}$(or $I_{T \rightarrow S}$), and train it similar to $G_{S \rightarrow T}$(or $I_{S \rightarrow T}$). 
At the first input frame, $x_{S,1}$, our framework uses an image generator, $I_{S \rightarrow T}$. Thus, we adopt 2D a cycle-consistency constraint~\cite{zhu2017unpaired} (see \figref{fig:mapping} (a)):

\begin{equation}
\begin{split}
L_{2Dcyc}(I_{S\rightarrow T}, I_{T\rightarrow S}) & {} \\
                                                = _{x_S \sim X_S} [ ||I_{T \rightarrow S}&(I_{S \rightarrow T}( x_{S,1} ) )-x_{S,1}||_1 ] \\
                                                + _{x_T \sim X_T} [ ||I_{S \rightarrow T}&(I_{T \rightarrow S}( x_{T,1} ) )-x_{T,1}||_1 ].
\label{eqn:cycle_loss_1}
\end{split}
\end{equation}

Given the previously generated frame, we now impose a new cycle-consistency constraint which regularizes the mapping across domains as well as over time (see \figref{fig:mapping} (c)):

\begin{equation}
\begin{split}
L_{3Dcyc}(G_{S\rightarrow T}, G_{T\rightarrow S}) & {} \\
                                                = _{x_S \sim X_S} [ ||G_{T \rightarrow S}&(G_{S \rightarrow T}( x_{S,t}, x_{S,t-1}, \hat{x}_{T,t-1} ) )-x_{S,t}||_1 ] \\
                                                + _{x_T \sim X_T} [ ||G_{S \rightarrow T}&(G_{T \rightarrow S}( x_{T,t}, x_{T,t-1}, \hat{x}_{S,t-1} ) )-x_{T,t}||_1 ].
\label{eqn:cycle_loss_2}
\end{split}
\end{equation}
For simplicity, note that ($\hat{x}_{T,t-1}, x_{S,t-1}$) and ($\hat{x}_{S,t-1}, x_{T,t-1}$) are omitted in the backward mapping functions, $G_{T \rightarrow S}$ and $G_{S \rightarrow T}$, respectively.
This loss essentially imposes a spatio-temporal constraint on the forward/backward mappings so that $G_{T \rightarrow S}(G_{S \rightarrow T}(\cdot)) \simeq x_{S}$, and similarly, $G_{S \rightarrow T}(G_{T \rightarrow S}(\cdot)) \simeq x_{T}$.

We compare our approach with existing CycleGAN-based methods in ~\figref{fig:mapping}. 
For the forward mapping, we use optical flow extracted from the input frames to warp and combine the previous output frame with the current generated frame through the fusion block.
The intermediate output, $\hat{x}_{T,t}$, is then mapped backward symmetrically with the forward mapping. Here, we use the same flow used in the forward mapping. We enforce cycle consistency on the final output. To the best of our knowledge, our approach is the first use-case of optical flow in an unsupervised video translation setting. Our experiments show that the warping enables the re-use of the pixels from the previous output, and this yields better video results than the RecycleGAN~\cite{bansal2018recycle} method which directly predicts all pixels / frames from scratch.

\subsection{Content Preserving Loss}

The original contents (\ie semantic labels) in a video tend to be mistranslated by existing CycleGAN-based frameworks~\cite{zhu2017unpaired,bansal2018recycle}. Indeed, conventional cycle-consistency does not necessarily guarantee the translation to be semantically consistent. This is because it does not consider any semantic correspondence during the translation, and thus the system can achieve perfect cycle-consistency (\ie, $L_{CYC}=0$) only if the inverse mapping recovers the original contents, regardless of how incorrect the forward mapping was.

In order to mitigate this label flipping problem, we adopt the content preservation constraint~\cite{shrivastava2017learning,taigman2016unsupervised,hoffman2017cycada}. Given a content extractor (e.g., VGGNet), we minimize the difference between the content of the original frame and its translation. We also apply instance normalization~\cite{ulyanov2016instance,huang2017arbitrary} before computing the feature distance, in order to wash out the style components, which leaves only content.

\begin{equation}
\begin{split}
L_{cont}(G_{S\rightarrow T}, G_{T\rightarrow S}, VGG) & {} \\
                                                     = || IN(VGG(G_{S \rightarrow T}( x_{S,t}, x_{S,t-1}, \hat{x}_{T,t-1} ))) - IN(VGG( x_{S,t} )) ||_2 \\
                                                     + || IN(VGG(G_{T \rightarrow S}( x_{T,t}, x_{T,t-1}, \hat{x}_{S,t-1} ))) - IN(VGG( x_{T,t} )) ||_2.
\label{eqn:semantic_loss}
\end{split}
\end{equation}

$IN(\cdot)$  and $VGG(\cdot)$ indicate instance normalization~\cite{ulyanov2016instance,huang2017arbitrary} and VGGNet respectively. We take the feature map of the 5\_3 th layer of VGG. By adding the above loss, we encourage the video frames to have the same semantics before and after the translations. Note that, because our formulation neither need the source domain label to exist nor use any of them during training, it is a more relaxed and general formulation compared to the previously proposed semantic loss in CyCADA~\cite{hoffman2017cycada}. We empirically verify its effectiveness in the experimental section. We omitted the content preserving loss of $I$ for clarity.

\subsection{Temporal Consistency Loss}

To reduce temporal flickering artifacts and false discontinuities in the video results, we employ a recurrence stream in the generator and train our model with flow warping loss. 
Note that, temporal dynamics of the translated videos should resemble those of the source videos. Thus, the temporal consistency loss of $L_{temp}$ is defined as

\begin{equation}
L_{temp} = \sum\limits_{t=2}^K O_{t \Rightarrow t-1} \left \| \hat{x}_{T,t} - W_{t \Rightarrow t-1}(\hat{x}_{T,t-1})\right\|_{1},
\label{eqn:tc_loss}
\end{equation}
where $O_{t \Rightarrow t-1}=\exp(-\alpha||x_{S,t}-W({x}_{S,t-1})||_2)$
denotes the occlusion mask which is calculated from the warping error between the input source frame $x_{S,t}$ and the warped source frame $W({x}_{S,t-1})$. 
Here, we set $\alpha=50$. The $W$ denotes the flow warping operation. We extract the backward flow from consecutive source frames $x_{S,t-1}, x_{S,t}$ using FlowNet2~\cite{ilg2017flownet} on-the-fly during training. We use the bilinear sampling layer to warp frames and set the number of recurrences to 3 $(K=3)$.

\subsection{Full objective}

Taking all the loss terms, our total loss function is as follows:

\begin{equation}
\begin{split}
L & =  L_{gan,forward} + L_{gan,backward} \\
  + & \lambda_{cyc}(L_{2Dcyc} + L_{3Dcyc})  + \lambda_{cont}L_{cont} + \lambda_{temp}L_{temp},
\label{eqn:final_loss}
\end{split}
\end{equation}
where the $\lambda_{cyc}$, $\lambda_{cont}$, $\lambda_{temp}$ denote weighting coefficients. The total loss objective enforces both semantic consistency and temporal consistency in the video results and thus is robust to semantic label flipping problem and temporal flicker artifacts.

\subsection{Training}

During training, the weighting coefficients
$\lambda_{cyc}$,
$\lambda_{cont}$
,and $\lambda_{temp}$ are set to 10, 1, and 10 respectively.
The resolution of the frames for all the experiments are set to $256 \times 512$. We adopt temporal flipping for data augmentation.
During training, we randomly retrieve 3 consecutive frames from the training videos to compute the temporal warping loss. 
We use the same network architecture for $G_{s \rightarrow t}$, and $G_{t \rightarrow s}$. The discriminator networks consist of a $70 \times 70$ PatchGAN ~\cite{isola2017image} that is used to classify $70 \times 70$ image pixels as real or fake.

\subsection{Testing}

At the first frame, we use an image generator to produce the translated result. This output is then fed into the recurrent generator by a feedback connection so that the generation is coherent with the past prediction.

\section{Experiments}

We use the VIPER dataset~\cite{Richter_2017} extracted from the game Grand Theft Auto V as our source domain. As the target domain, we use the Cityscapes dataset, from which we used 2975 videos, each with 30 frames. Therefore, we train our model to translate a video from VIPER to that of the Cityscapes dataset. For some semantic categories, that are only present in Cityscapes but not in the VIPER dataset,~\eg \textit{wall} and \textit{rider}, we ignore these classes in the label transfer for evaluation. We compare our method with the state-of-the-art baseline methods both quantitatively and qualitatively. Finally, we demonstrate the applicability of our framework on the domain adaption setting.

\paragraph{Baselines.} We compare our approach to the following five baselines. For the per-frame methods~\cite{zhu2017unpaired,hoffman2017cycada}, we build their temporally extended versions by running a post processing method~\cite{lai2018learning} on their video results. We run their test codes with our testing videos.
\begin{itemize}
\item CycleGAN~\cite{zhu2017unpaired}: The state-of-the-art unpaired image-to-image translation method. We run their method in a frame-by-frame manner to perform video translation.
\item CyCADA~\cite{hoffman2017cycada}: They introduce semantic loss in translation for image-based domain adaptation. Unlike their original implementation, we include semantic loss to improve their semantic segmentation performance to build this baseline. This baseline is run in a frame-by-frame manner.
\item RecycleGAN~\cite{bansal2018recycle}: The state-of-the-art unpaired video-to-video translation approach.
\item CycleGAN~\cite{zhu2017unpaired} + Blind Temporal Consistency(BT)~\cite{lai2018learning}: We post-process CycleGAN results for temporal consistency.
\item CyCADA~\cite{hoffman2017cycada} + Blind Temporal Consistency(BT)~\cite{lai2018learning}: We post-process CycleGAN results for temporal consistency.
\end{itemize}

\subsection{Qualitative Results}
\begin{figure*}
\begin{tabular}{@{}c@{}}
\includegraphics[width=0.99\linewidth]{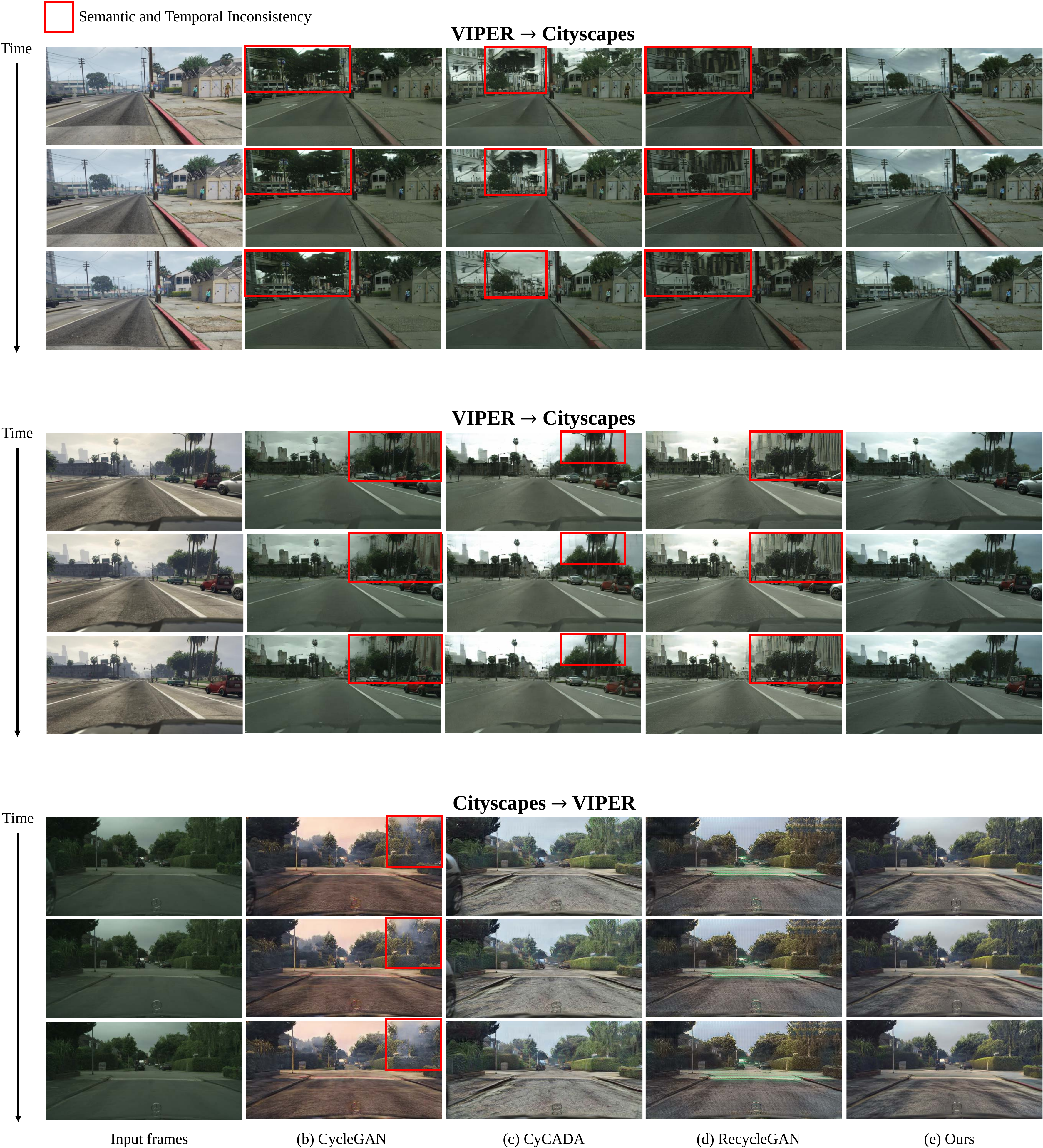} \\
\end{tabular}
\caption{ {\bf Qualitative Results.} We sampled the translated video frames consecutively. We indicate red box for semantic and temporal inconsistency of video frames. We can see that previous approaches suffer from both the semantic label flipping problem and the temporal inconsistent generation. Instead, our framework successfully resolves both issues. \textit{Best viewed in color}.}
\label{fig:qual}
\end{figure*}

In \figref{fig:qual}, we show three examples (\ie~examples of two forward and one backward mapping) to visually compare our video results with the baseline methods. We sampled the translated video frames consecutively. We include red boxes when there was semantic and temporal inconsistency in the video frames. 

We can observe that the previous methods all suffer from both temporal flicker artifacts and semantic label flipping problem. Compared to the previous approaches, our framework effectively reduces both issues.

\begin{figure*}
\begin{tabular}{@{}c@{}}
\includegraphics[width=0.90\linewidth]{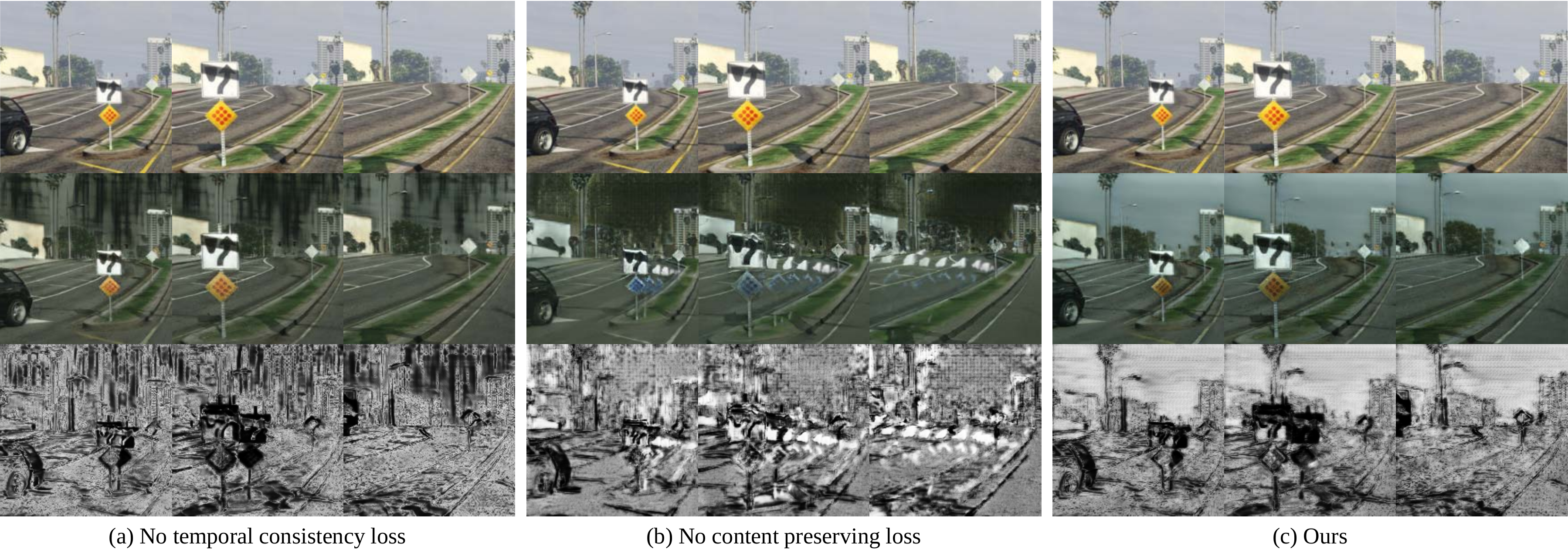}\\
\end{tabular}
\vspace{-4mm}
\caption{ {\bf Ablation studies.} Each row shows three input frames, translated frames, and corresponding fusion masks respectively. 
\textit{Best viewed in color}.}
\label{fig:qual_ab}
\end{figure*}

\subsection{Semantic Consistency}
We adopt the same evaluation protocol as the previous work ~\cite{wang2018high}, which suggested that a well-translated frame from source to target domain will work as an indistinguishable input for a network that has been trained on target domain data. More specifically, if the video translation from VIPER to Cityscapes was successful, the semantic segmentation network that has been trained on the Cityscapes dataset will perform well on the translated frames. We use three metrics to evaluate the semantic segmentation performance: mean intersection-over-union (mIOU), frequency weighted intersection-over-union (fwIoU), and pixel accuracy.
We use an off-the-shelf Cityscapes-pretrained segmenter, PSPNet~\cite{zhao2017pyramid}, in this experiment. Our model and all other baselines share the same training dataset, and the pretrained PSPNet is run on their translation results of test set. As shown in~\tabref{tab:segmentation}, our model outperforms all baselines in all evaluation metrics. This indicates that our method translates the source frames closest toward the distribution of the target domain,~\ie Cityscapes, while maintaining the best semantic consistency during translation.


\subsection{Temporal Consistency}

We investigate the temporal consistency of the video results. In practice, we measure the temporal error of a translated video sequence, which is the average pixel-wise Euclidean color difference between consecutive frames. We use groundtruth optical flows in the VIPER dataset to warp the images. Table~\ref{tab:warping} shows that the temporal error is significantly reduced compared to the other methods. Note that enforcing temporal consistency in post-processing~\cite{lai2018learning} only reduces the temporal warping error and does not ensure semantic consistency (see ~\tabref{tab:segmentation}). On the contrary, because our method spatio-temporally regularizes the translation all at once, the results are consistent both semantically and temporally, showing the best results in ~\tabref{tab:segmentation} and ~\tabref{tab:warping}.

\begin{table}[t]
\centering
\begin{tabular}{ l|c c c }
\hline
\hline
&\rotatebox[origin=c]{0}{mIOU} & \rotatebox[origin=c]{0}{fwIOU} &\rotatebox[origin=c]{0}{Pixel acc.}\\
\hline
CycleGAN~\cite{zhu2017unpaired} & 25.54 & 49.73 & 65.07 \\ 
CycleGAN~\cite{zhu2017unpaired} +BT~\cite{lai2018learning} & 26.07 & 49.41& 64.87\\ 
CyCADA~\cite{hoffman2017cycada} & 30.75 & 58.81 & 72.71\\ 
CyCADA~\cite{hoffman2017cycada} +BT~\cite{lai2018learning} & 30.58 & 57.04& 71.38\\ 
RecycleGAN~\cite{bansal2018recycle} & 25.10 & 48.85& 64.02\\ 
Ours & \textbf{35.14} & \textbf{65.08} & \textbf{77.13}\\ 
\hline
PSPNet on Cityscapes &64.69 &86.27 &92.18\\
\hline
\hline
\end{tabular}
\caption{\textbf{Semantic segmentation scores on 512 $\times$ 1024 resolution.} }
\vspace{1mm}
\label{tab:segmentation}
\end{table}

\begin{table}[t]
\centering
\begin{tabular}{ l|c }
\hline
\hline
&\rotatebox[origin=c]{0}{Warping error.}\\
\hline
CycleGAN~\cite{zhu2017unpaired} & 0.002017 \\
CycleGAN~\cite{zhu2017unpaired}+BT~\cite{lai2018learning} & 0.000829 \\
CyCADA~\cite{hoffman2017cycada} & 0.002003 \\
CyCADA~\cite{hoffman2017cycada}+BT~\cite{lai2018learning} & 0.000945 \\
RecycleGAN~\cite{bansal2018recycle} & 0.001640  \\
Ours & \textbf{0.000437} \\
\hline
Oracle - Source (VIPER)                  & 0.000264 \\
\hline
\hline
\end{tabular}
\caption{\textbf{Flow warping errors.} We evaluate the flow warping errors of target-like source videos. We use ground-truth optical flow from the VIPER dataset.}
\vspace{1mm}
\label{tab:warping}
\end{table}

\begin{table}
\centering
\begin{tabular}{ l|c c c }
\hline
\hline
&\rotatebox[origin=c]{0}{mIOU}  &\rotatebox[origin=c]{0}{Pixel acc.} &\rotatebox[origin=c]{0}{Warping error.}\\
\hline
Ours & \textbf{35.14} & \textbf{77.13} & \textbf{0.000437}\\
(-) temporal consistency  & 29.58 & 67.86 & 0.001241\\
(-) content preserving  & 22.09 & 60.04 & 0.001082 \\
\hline
\hline
\end{tabular}
\vspace{1mm}
\caption{\textbf{Ablation studies.} We compute the semantic segmentation score and the flow warping loss to quantitatively determine the effects of the proposed loss functions. }
\label{tab:ablation}
\end{table}

\begin{table*}[t]
\centering
\resizebox{\textwidth}{!}{\begin{tabular}{l c c c c c c c c c c c c c c c c c c c c }
\hline 
\hline
\multicolumn{21}{c}{VIPER $\rightarrow$  Cityscapes} \\
\hline
\\
Method &\rotatebox[origin=l]{90}{road} &\rotatebox[origin=l]{90}{sidewalk} & \rotatebox[origin=l]{90}{building} &\rotatebox[origin=l]{90}{wall} &\rotatebox[origin=l]{90}{fence} &\rotatebox[origin=l]{90}{pole} &\rotatebox[origin=l]{90}{traffic light}& \rotatebox[origin=l]{90}{traffic sign} &\rotatebox[origin=l]{90}{vegetation} &\rotatebox[origin=l]{90}{terrain} &\rotatebox[origin=l]{90}{sky} &\rotatebox[origin=l]{90}{person} &\rotatebox[origin=l]{90}{rider}&\rotatebox[origin=l]{90}{car}&\rotatebox[origin=l]{90}{truck}&\rotatebox[origin=l]{90}{bus}&\rotatebox[origin=l]{90}{train}&\rotatebox[origin=l]{90}{motorbike}&\rotatebox[origin=l]{90}{bicycle}    &\rotatebox[origin=l]{90}{mIOU} \\
\hline
Source only & 6.6  & 3.0 & 44.7 &	0 & 4.9 &	12.7 & 10.6 & 3.6 & 67.0 & 6.1 & 67.8 &	3.7 &	0 &	14.8 & 0.6 & 0.3 & 0 &0 &0 &13.0 \\
CycleGAN~\cite{zhu2017unpaired}  & 81.6  & 18.2 & 59.8&	0 & 1.0 & 11.3 &8.1 & 2.8& 76.4& 16.7& 52.3& 2.9&  0  & 51.2 &  4.7   & 1.5&0	 &0 &0  & 20.5\\
RecycleGAN~\cite{bansal2018recycle} & 82.3  &10.2 &64.9 &0	 &1.1 & 14.4 &12.0 &1.9 &73.5 &20.3  &69.2 & 3.8 &  0  & 44.8 & 4.9  &0.3 &0  & 0 &   0   & 21.2 \\
CyCADA~\cite{hoffman2017cycada} &  84.1 & 22.8& 65.9&0 & 1.5 & 17.0 & 7.4& 2.2& 74.2& 19.8&58.5 & 3.1 &   0  & 50.3 & 3.2  & 6.0 & 0 & 0&0  & 21.9 \\
Ours    & 85.8 & 23.9& 65.1& 0& 5.6 & 9.4 & 15.0& 2.6& 76.2 & 26.7& 76.0&8.8	&0	 &54.9	&6.5	 & 1.0 & 0& 0 &0 & \textbf{24.1} \\
\hline
Oracle - Target  & 94.5	& 65.9& 84.5& 17.3& 27.5& 45.3& 43.8& 56.3& 86.8&38.7 & 87.3&65.8 & 40.4&	79.5&	19.2 & 15.5& 3.9& 21.8& 59.9&50.2\\
\hline
\hline
\end{tabular}}
\caption{{\textbf{Domain adaptation results.} We compared the proposed method to }CycleGAN~\cite{zhu2017unpaired}, CyCADA~\cite{hoffman2017cycada}, and RecycleGAN~\cite{bansal2018recycle}. We also experimented using only the source, as well as having data from the target domain.}
\label{tab:domain_adoption}
\vspace{-5mm}
\end{table*}

\subsection{Ablation studies}

We run ablation studies to demonstrate the effects of the two proposed objective functions: 1) content preserving loss and 2) temporal consistency loss. We compute semantic segmentation scores and flow warping errors. To evaluate the setup, we zero out the loss functions during training. We summarize the results in ~\tabref{tab:ablation}. In both ablation setups, we observe significant performance drops. The low scores of these setups show that two objective functions are all important for learning a good translation. Moreover, we show each components' effects qualitatively in ~\figref{fig:qual_ab}. We observe that the model without the temporal consistency loss misses to utilize temporal information during training and generates temporally unstable videos (e.g., the background content varies in ~\figref{fig:qual_ab} (a)).
On the other hand, ablating the content preserving loss instead lead to significant smearing artifacts (~\figref{fig:qual_ab} (b)). This implies that the content preserving loss plays an important role in compensating the intrinsic error in the flow~\cite{ilg2017flownet}. We clearly observe that using both objective functions produces the best translated result. Note that wrongly warped pixels (e.g., the traffic sign) are successfully replaced with synthesized pixels in the fusion block.

\subsection{User study results}
We perform a user study to evaluate the visual quality of the translated videos. We compare our method with the state-of-the-art baselines. We use a total of 30 translated videos: 15 videos from the forward mapping (VIPER $\rightarrow$ Cityscapes) and the rest from the backward mapping (Cityscapes $\rightarrow$ VIPER). In each testing case, we first show the original input video, and then our result and the other results on the same screen. The order of the video results is shuffled. To ensure that a user has enough time to distinguish the difference and make a careful judgement, we allow users to see the videos several times. Each participant is asked to choose their most preferred result. A total of 30 users participated in this study. We specifically ask each participant to check for both semantic and temporal consistency. The user study results are summarized in \tabref{tab:user_study}. It shows that our method is most preferred by the participants. The complete video results can be found in the supplementary materials. 

\begin{table}[t]
\centering
\begin{tabular}{ l|c c c }
\hline
\hline
\multicolumn{3}{c}{Q. Which one do you prefer? ($\%$)} \\
\hline
&\rotatebox[origin=c]{0}{forward}  &\rotatebox[origin=c]{0}{backward}\\
\hline
CycleGAN~\cite{zhu2017unpaired} & 0.22 & 11.33 \\
CycleGAN~\cite{zhu2017unpaired}+BT~\cite{lai2018learning} & 0.89 & 18.0 \\
CyCADA~\cite{hoffman2017cycada} & 10.44 & 6.0 \\
CyCADA~\cite{hoffman2017cycada}+BT~\cite{lai2018learning} & 4.67 & 14.0 \\
RecycleGAN~\cite{bansal2018recycle} & 4.0 & 8.89  \\
Ours & \textbf{79.78} & \textbf{41.78} \\
\hline
\hline
\end{tabular}
\caption{\textbf{Comparisons of user preference.} forward and backward indicate VIPER $\rightarrow$  Cityscapes and Cityscapes $\rightarrow$ VIPER respectively. }
\label{tab:user_study}
\vspace{-5mm}
\end{table}

\subsection{Domain Adaptation: Segmentation}
We evaluate our model in a synthetic (VIPER) to real (Cityscapes) adaptation setting. The task is to predict semantic label maps. The results are summarized in Table~\ref{tab:domain_adoption}.

\sloppy  Specifically, we first translate VIPER frames into Cityscapes-like frames using the trained generator. The semantic segmentation model~\cite{Yu2017} is then trained using the translated frames and semantic labels available in the VIPER dataset.
Our \textit{vanila} baseline is trained on source (VIPER) frames without any domain adaptation, and suffers a huge performance drop. On the other hand, we observe that our translation method indeed narrowed the domain gap, so that the segmentation model trained on our translated frames outperforms all the competitors by a large margin. We emphasize that while the CyCADA (semantic loss) requires semantic labels from the source domain to train their translation model, our method performs better without relying on such extra label data for both 
training and testing.

\section{Conclusion}
We propose a novel framework for unpaired video-to-video translation. Our framework enforces both semantic and temporal consistency in the video results and thus is robust to semantic label flipping problem and temporal flicker artifacts. Our video translation is trained to have correct semantic correspondences between the source and target domains at patch level, which cannot be guaranteed with the existing cycle/recycle consistency loss. Also, our recurrent generator, coupled with temporal loss leads to temporally smooth video results, because the generation at each time step is encouraged to be coherent with the past predictions. Finally, we show that our framework also can achieve favorable results in the domain adaptation task. We hope our proposed framework becomes an important architecture for solving real-world video translation problems.

%
\begin{acks}
This research was supported in part by Samsung Electronics.
\end{acks}

%

\bibliographystyle{ACM-Reference-Format}
\balance
\bibliography{egbib}

%
\clearpage
\appendix

\section{Appendix}

\subsection{Implementation details}

Our model is implemented using Pytorch v0.4, CUDNN v7.0, CUDA v9.0. It runs at 14.3 fps on hardware with Intel(R)Xeon(R) (2.10GHz) CPU and NVIDIA GTX 1080 Ti GPU.
For training, we adopt Adam optimizer with $\beta$= (0.5, 0.999) and a fixed learning rate of 2e-4. 
The overall training takes 2 days on one NVIDIA GTX 1080 Ti GPU.

\textbf{The image generator}($I_{s \rightarrow t}$, $I_{t \rightarrow s}$) architecture follows ~\cite{zhu2017unpaired}, using six residual blocks.
We use the same network architecture for $G_{s \rightarrow t}$, and $G_{t \rightarrow s}$. The discriminator networks consist of a $70 \times 70$ PatchGAN that is used to classify $70 \times 70$ image pixels as real or fake.

\textbf{The flow estimator} architecture follows ~\cite{ilg2017flownet}. Thus, two inputs are concatenated.
If we increase the content loss weight to become high, we observe that the model tends to fall into a trivial solution (identity mapping).
The warping operation is implemented via a simple bilinear interpolation based on the estimated flow field ~\cite{ilg2017flownet}.

\textbf{The fusion block} is composed of 3 convolution layers. The InstanceNorm and the ReLU operations are followed except the last layer. The sigmoid layer is used at the end to constrain the values of the mask. To ensure that the output size would be equal to the input feature map, we use reflection padding.

\subsection{More ablation studies}
\textbf{Ablation studies on fusion block} We investigate the effectiveness of learning a soft fusion mask.
We compare our method with two simple baselines : 1) an averaging and 2) a rule-based occlusion mask (\eqnref{eqn:tc_loss}).
For each method, we train the model from scratch. The results are summarized in ~\tabref{tab:ablation2}.
We clearly observe that those baselines produce scores far below compared to our method.
Specifically, in the case of averaging, the method propagates the wrongly warped pixels, rather than properly attenuating them. Thus, the output videos show significant smearing artifacts.
For the rule-based occlusion mask, we see that the model fails to converge during training, giving the worst scores.
On the other hand, our fusion block identifies the wrong flow well and properly takes the currently synthesized pixels to compensate errors (see Figure 4).

\begin{table}[h]
\centering
\begin{tabular}{ l|c c c }
\hline
\hline
&\rotatebox[origin=c]{0}{mIOU}  &\rotatebox[origin=c]{0}{Pixel acc.} &\rotatebox[origin=c]{0}{Warping error.}\\
\hline
Ours & \textbf{35.14} & \textbf{77.13} & \textbf{0.000437}\\
Average & 28.78 & 69.38 & 0.000649\\
Rule based mask& 15.65 & 54.58 & 0.000778\\
\hline
\hline
\end{tabular}
\caption{\textbf{Ablation studies on fusion block.} }
\label{tab:ablation2}
\end{table}

\subsection{Experiments on other datasets}
To show the generality of the proposed framework, we further apply our method to two different datasets (face-to-face and flower-to-flower) and compare the results with the state-of-the-art unpaired video translation method, RecycleGAN~\cite{bansal2018recycle}.

\noindent \textbf{Face to Face}: We train our model using the videos of various public people(e.g., John Oliver, Stephen Colbert, Barack Obama and Donald Trump). The videos are cropped around the face region based on the facial keypoints. 
We observe that both methods learn a plausible translation.

Compared to the image based translation method~\cite{zhu2017unpaired}, which suffers from severe mode collapse when translating the videos, ours and the RecycleGAN method generate appealing results. This is because both exploit temporal information during translation.

\noindent \textbf{Flower to Flower}: We also train our model using flower dataset. We observe tendencies similar to those shown in the face to face case.

We argue that both methods produce good results because each set has trivial semantic patterns and easy movements. However, we show that our method significantly outperforms RecycleGAN~\cite{bansal2018recycle} when the translation scenario becomes very challenging: in the synthetic (VIPER) to real (Cityscapes) driving scene video translation.




\end{document}